\pdfoutput=1

\documentclass[11pt]{article}

\usepackage{acl}

\usepackage{times}
\usepackage{latexsym}

\usepackage[T1]{fontenc}

\usepackage[utf8]{inputenc}

\usepackage{microtype}

\usepackage{graphicx}
\usepackage{tabularx}
\usepackage{soul}

\usepackage{epstopdf}
\usepackage[utf8]{inputenc}

\usepackage{hyperref}
\usepackage{xstring}

\usepackage{color}
\usepackage{multirow}

\usepackage{multibib}

\usepackage{graphicx}
\usepackage{amsmath}
\usepackage{amsthm}
\usepackage{booktabs}
\usepackage{algorithm}
\usepackage{algorithmic}
\usepackage{amssymb}
\usepackage{xcolor}

\title{The JDDC 2.0 Corpus: A Large-Scale Multimodal Multi-Turn Chinese Dialogue Dataset for E-commerce Customer Service}

\author{Nan Zhao, Haoran Li, Youzheng Wu, Xiaodong He, Bowen Zhou \\ \\
	 JD AI Research\\
\texttt{\{zhaonan8,lihaoran24,wuyouzheng1,hexiaodong,zhoubowen\}@jd.com}\\
}

\begin{document}
\maketitle
\begin{abstract}
With the development of the Internet, more and more people get accustomed to online shopping. When communicating with customer service, users may express their requirements by means of text, images, and videos, which precipitates the need for understanding these multimodal information for automatic customer service systems.
Images usually act as discriminators for product models, or indicators of product failures, which play important roles in the E-commerce scenario.
On the other hand, detailed information provided by the images is limited, and typically, customer service systems cannot understand the intents of users without the input text.
Thus, bridging the gap of the image and text is crucial for the multimodal dialogue task.
To handle this problem, we construct JDDC 2.0, a large-scale multimodal multi-turn dialogue dataset collected from a mainstream Chinese E-commerce platform\footnote{\url{https://JD.com}}, containing about 246 thousand dialogue sessions, 3 million utterances, and 507 thousand images, along with product knowledge bases and image category annotations. 
We present the solutions of top-5 teams participating in the JDDC multimodal dialogue challenge based on this dataset, which provides valuable insights for further researches on the multimodal dialogue task.

\end{abstract}

\section{Introduction}

\begin{figure*}[!htb]
	\centering
	\includegraphics[width=5 in]{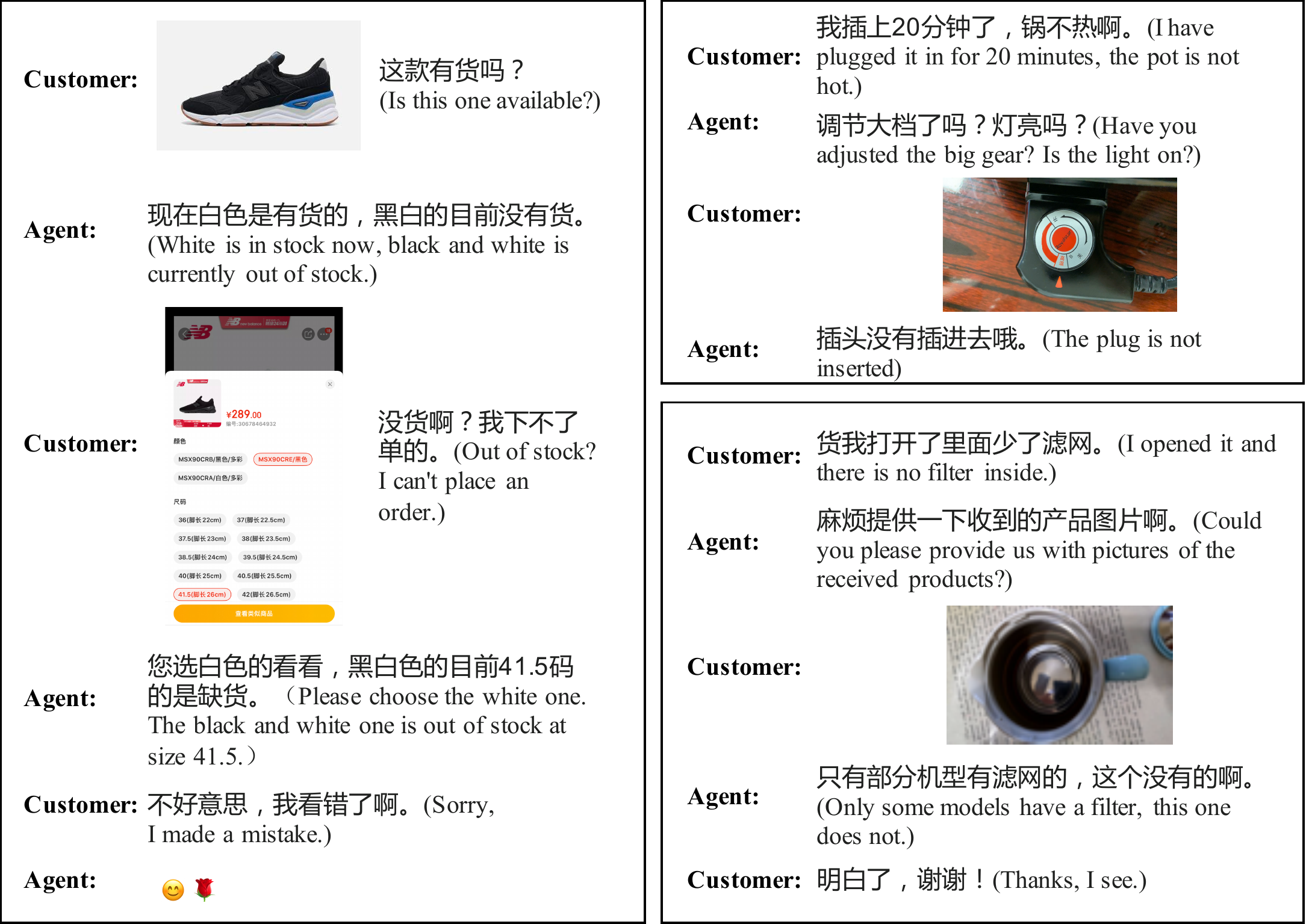}\\
\caption{Three segments of dialogue sampled from our JDDC 2.0 corpus.
The responses cannot be generated only from neither images nor texts, showing the necessity of jointly understanding the textual and visual information in dialogues. }
\label{fig1}
\end{figure*}

In the era of the Internet, online shopping has brought tremendous convenience to people.
When users encounter difficulties in the process of online shopping, they would contact the customer service for helps.
There are various forms of information when users communicating with customer services, including text, images, or even videos, which establishes a great challenge for customer service systems to understand users' requirements.
Generally, users tend to express their needs with text.
While sometimes, text fails to convey enough information, and in that case, users may resort to images.
For example, in Figure~\ref{fig1}, images are used for distinguishing different product models for the same brand, or used for identifying the location and cause of product failures.
Therefore, there is an urgent need for customer service systems to understand multimodal information sent by users to provide proper responses.

Multimodal information processing has been widely explored recently. Researches on image captioning~\cite{xu-15,Anderson_2018_CVPR,pan2020x}, visual question answering~\cite{Antol-15,Yang_2016_CVPR,lu2016hierarchical},
multimodal machine translation~\cite{calixto-etal-2017-doubly,caglayan-etal-2017-lium,helcl-etal-2018-cuni},
multimodal summarization~\cite{li-etal-2017-multi,ijcai2018-0577,li2020aspect,li-etal-2020-multimodal},
and visual dialogue~\cite{visdia-17,das2017learning,murahari2020large} have made remarkable progress.
However, the application of multimodal dialogue in E-commerce scenarios remains to be studied.
In this paper, we build a large-scale multimodal dialogue dataset in E-commerce that aims to boost the research on the multimodal dialogue task.

Early work~\cite{tw-11,weibo-13,tw-15,ubuntu-15,per-16,rd-18} construct dialogue corpus with discussion records on the social media, such as Twitter, Sina Weibo, and Reddit.
Although the discussions on social media consist of multiple turns, they are quite different from real dialogues due to the lack of an explicit goal in the conversation~\cite{mul-18}.
To mitigate this problem, researches~\cite{dcc-18,mul-18,pers-18,wiki-19,risa-20} build the dialogue datasets through crowd-sourcing by asking annotators talk to each other according to given dialogue objectives.
While in the real world, topic switches across multi-domain and emotional interactions are more abundant, and thus,  dialogue datasets in real scenario are valuable for researches.
The JDDC corpus (Jing Dong Dialogue Corpus)~\cite{Chen-20} is a dialogue dataset consisting of conversations about after-sales topics in E-commerce scenario, which is goal-driven and with long-term dependency among the context.
In addition, JDDC contains task-oriented, chitchat, and question-answering dialogues.

With the widespread use of smartphones, taking screenshots and photos have been very convenient, 
and users often describe their issues by a combination of text and image when they communicate with customer service,
which motivates us to explore the multimodal dialogue task.
In this paper, we construct the JDDC 2.0 corpus\footnote{Dataset is now available at \url{https://jddc.jd.com}. Users need to register and require authorization to obtain the data.} that is composed with multimodal dialogues where each dialogue session contains multiple pieces of text and at least one image.
Other characteristics of the previous JDDC Corpus are maintained.
JDDC 2.0 contains about 246 thousand dialogue sessions, 3 million utterances, and 507 thousand images, along with product knowledge bases and image category annotations. 
In addition, we introduce the solutions of top-5 teams participating in the multimodal dialogue challenge with JDDC 2.0.

\section{Related Work}

Visual Dialogue dataset (VisDial)~\cite{visdia-17} first introduces visual contents into dialogues, in which the utterances are  a group of questions and answers towards the corresponding image. 
This dataset focuses on understanding the given images.
\citet{IGC-17} observes that, in social media, the information for conversations around images are beyond what is visible in the image. They propose a new task called image-grounded conversation (IGC) that aims to constitute conversations with the images as the grounding, where the objects in images may not be mentioned in the conversation. In other words, images in IGC act as the conversation topics. 
Similar to IGC, Image-Chat dataset~\cite{imgchat-20} is also an image-grounded dialogue dataset, where the dialogue is performed based on a given emotional mood or style, which are key factors in engagingness~\cite{guo2019mscap}.

\begin{table*}[!htb]
	\centering
	\begin{tabular}{lcccc}
	\toprule
	{Dataset} & {Language} & {\# Dialogues} & {\# Average Turns} & {\# Images}  \\
	\midrule
    VisDial~\cite{visdia-17}    & English & 120,000                  & 20                     & 120,000         \\
    IGC~\cite{IGC-17}              & English & 4,222                 & 6                      & 4,222              \\ 
    Image-Chat~\cite{imgchat-20}       & English & 201,779               & 2                      & 201,779            \\ 
    MMD~\cite{MMD-18}     & English & 150,629               & 40                     & 385,969            \\ 
    SIMMC 2.0~\cite{smmc-21}        & English & 11,244                & 10.40                   & 1,566             \\ 
    \midrule
    JDDC 2.0 (Ours)       & Chinese & 246,153               & 14.06                  & 507,678            \\
    \bottomrule
	\end{tabular}%
	\caption{Comparison of JDDC 2.0 with other multimodal dialogue datasets.}
	\label{tab.ds}
\end{table*}%

\begin{table*}[!htb]
\scalebox{0.85}{
\begin{tabular}{|l|c|c|c|c|c|c|c|}
\hline
\multirow{2}{*}{}           & \multirow{2}{*}{\textbf{All}} & \multicolumn{2}{c|}{\textbf{Training Set}}                                                                                                  & \multicolumn{2}{c|}{\textbf{Validation Set}}                                                                                               & \multicolumn{2}{c|}{\textbf{Test Set}}                                                                                                  \\ \cline{3-8} 
                            &                      & \begin{tabular}[c]{@{}c@{}}Home\\ appliances\end{tabular} & \begin{tabular}[c]{@{}c@{}}Fashion\end{tabular} & \begin{tabular}[c]{@{}c@{}}Home\\ appliances\\ \end{tabular} & \begin{tabular}[c]{@{}c@{}}Fashion\end{tabular} & \begin{tabular}[c]{@{}c@{}}Home\\ appliances\end{tabular} & \begin{tabular}[c]{@{}c@{}}Fashion\end{tabular} \\ \hline
\# Dialogues                & 246,153              & 103,555                                                               & 93,371                                                     & 12,941                                                                & 11,674                                                     & 12,935                                                                & 11,674                                                     \\ \hline
\# Utterances               & 3,459,888            & 1,481,151                                                             & 1,284,819                                                  & 185,879                                                               & 162,630                                                    & 185,197                                                               & 160,217                                                    \\ \hline
\# Avg.  turns per dialogue  & 14.06                & 14.30                                                                  & 13.76                                                      & 14.36                                                                 & 13.93                                                      & 14.32                                                                 & 13.72                                                      \\ \hline
\# Images                   & 507,678              & 210,386                                                               & 195,549                                                    & 26,218                                                                & 24,696                                                     & 26,031                                                                & 24,888                                                     \\ \hline
\# Avg. images per dialogue & 2.06                 & 2.03                                                                  & 2.09                                                       & 2.03                                                                  & 2.11                                                       & 2.01                                                                  & 2.13                                                       \\ \hline
Avg. conversation length    & 27.24                & 23.77                                                                 & 31.25                                                      & 23.81                                                                 & 31.09                                                      & 23.68                                                                 & 31.29                                                      \\ \hline
\end{tabular}
}
\caption{Statistics of the JDDC 2.0 dataset.}
\label{tab.pro}
\end{table*}

Multimodal dialogue is different from visual dialogue, and the characteristics of multimodal dialogue can be summarized as follows.
(1) There may be more than one image for a multimodal dialogue session. 
(2) The images can be updated with the advance of dialogue. 
(3) The questions and answers can be multimodal or monomodal. 
(4) Knowledge bases may be useful for multimodal dialogue.
(5) Multimodal dialogue models sometimes need to clarify users' requirements with some dialogue strategies like rhetorical questions.

As dialogue systems are widely deployed in the E-commerce domain, and nowadays, many online customer service robots have been put into use to provide 24-hour services to help customer solve various problems in the process of online shopping. 
There has been existing multimodal dialogue datasets in the E-commerce domain.
\citet{MMD-18} build the Multimodal Dialogs dataset (MMD) that consists of over 150K dialogue sessions between shoppers and sales agents, with 84 dialog states and various conversations flows suggested by fashion retail experts.
The dialogue scene in MMD is limited for the  pre-sales guidance, while other scenes, such as payment, logistics, and after-sales maintenance, are not covered.
The dataset of SIMMC 2.0~\cite{smmc-21} contains 11K dialogue sessions between costumers and virtual assistants for situated and photo-realistic VR applications.
Similar to MMD, the target scenes for SIMMC 2.0 is the pre-sales guidance.
In fact, changes for scene are quite frequent. For example,  a dialogue system usually need to solve customer's problem ranging from product selection, payment, to logistics and distribution in a dialogue session. 
Thus, in this paper, we collect the JDDC 2.0 dataset that covers almost the complete process in E-commerce.
Table~\ref{tab.ds} shows the detailed comparison of JDDC 2.0 with other existing multimodal dialogue datasets.

\section{The JDDC 2.0 Corpus}

\subsection{Dataset Collection}

We collect our dataset from \url{JD.com}, a large E-commerce platform in China.
We select the conversations between users and customer service for two categories of products of large sales volume, including small home appliances and fashion, as the source of our dataset. 
In the real E-commerce scenario, the type of dialogue is diversified, where the customer service needs to answer the questions posed by users passively and recommend products to users actively. 
To ensure a high quality in diversity, in the process of data selection, we only select conversations of customer service staff with gold medals, who tend to answer the questions more accurately and recommend products more suitably than general staff.
In addition, according to our observations, dialogue behaviors for customer service staff with gold medals are richer and more natural than general ones.
We collect conversation logs of one month and finally maintain the conversations containing at least one images as our dataset. 

\subsection{Dataset Statistics}

Table~\ref{tab.pro} shows the statistics of JDDC 2.0 that contains 246,153 dialogue sessions, 3,459,888 utterances, and 507,678 images. The dataset is divided into the training set, the validation set, and the test set according to the ratio of 80\%, 10\%, and 10\%.
The number of dialogues in the two categories is roughly equal.

\begin{figure}[!htb]
\centering
\includegraphics[width=3 in]{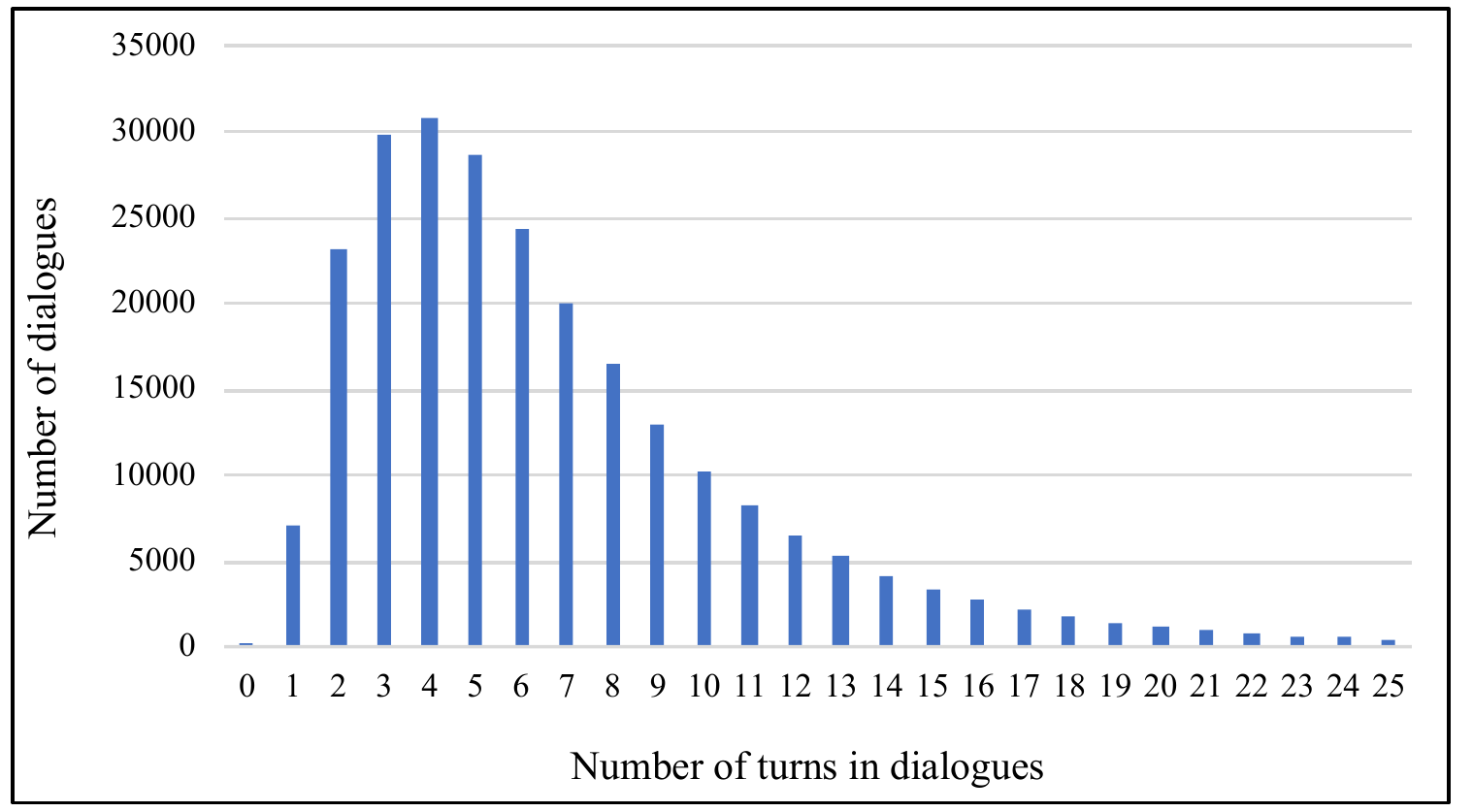}\\
\caption{The statistics of dialogue turns.}
\label{fig2}
\end{figure}

\begin{figure}[!htb]
\centering
\includegraphics[width=3 in]{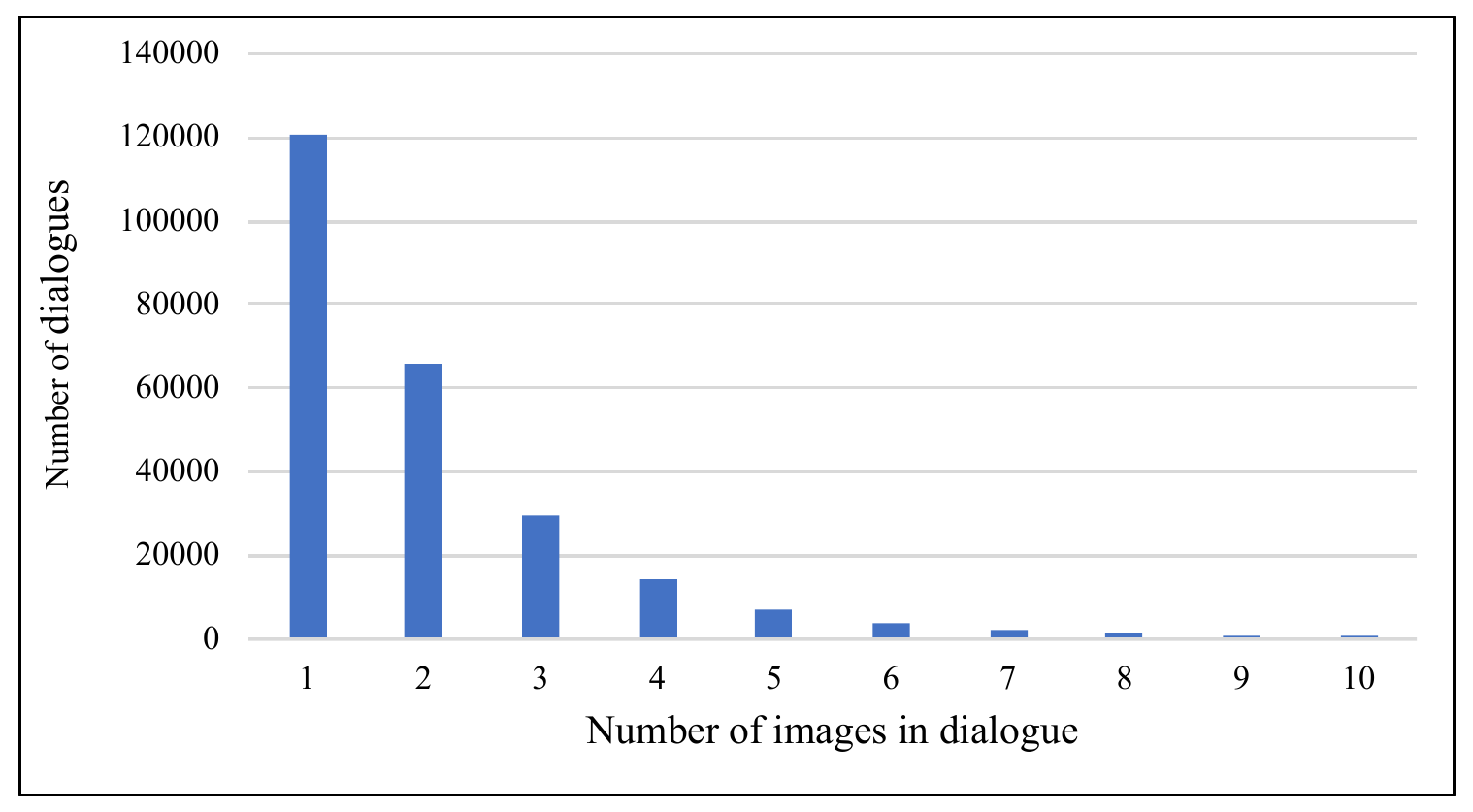}\\
\caption{The statistics of numbers of image.}
\label{fig3}
\end{figure}

\begin{figure}[!htb]
\centering
\includegraphics[width=3 in]{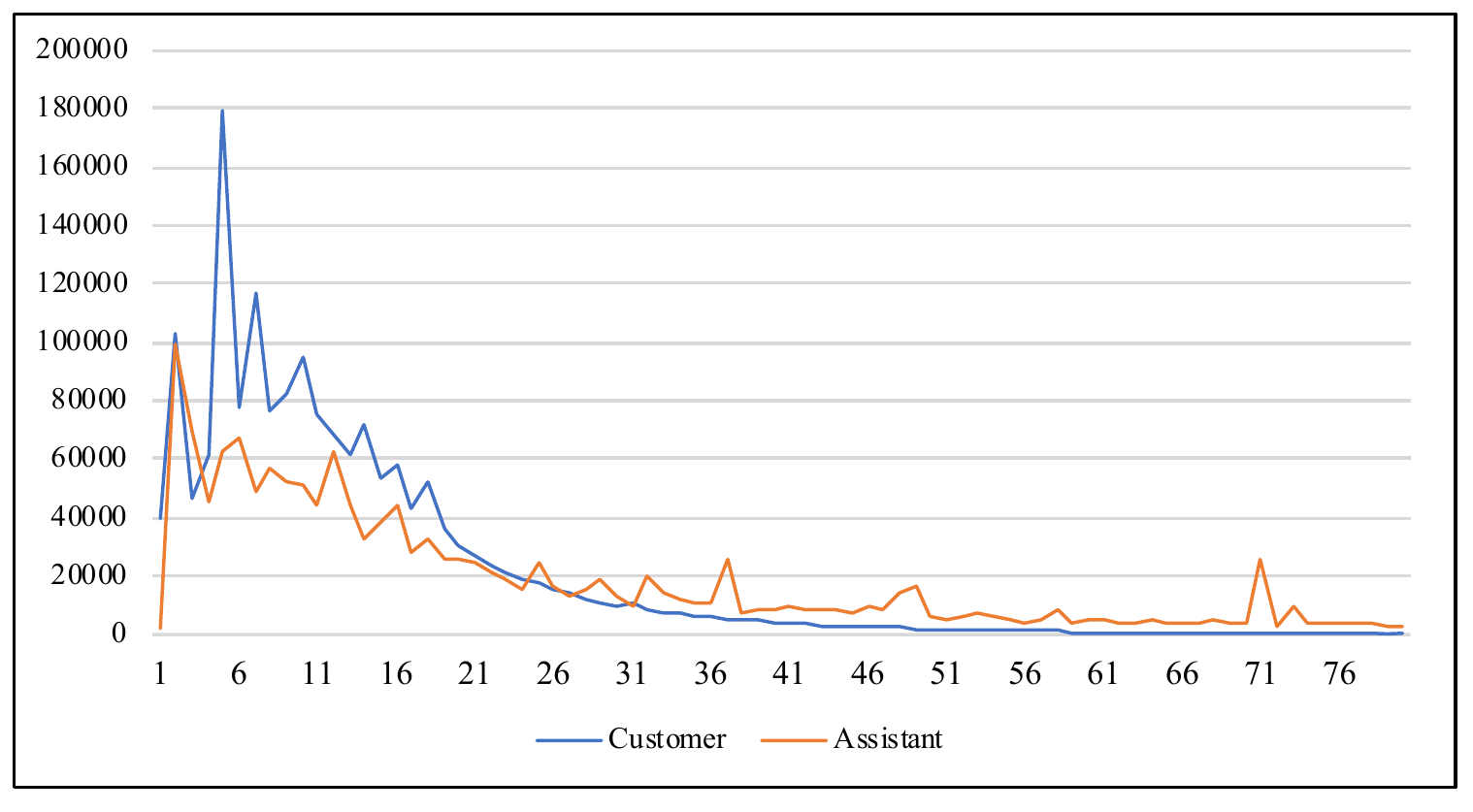}\\
\caption{The statistics of dialogue utterance length.}
\label{fig4}
\end{figure}

Figure~\ref{fig2} demonstrates the statistics of dialogue turns.
The average number of dialogue turns in JDDC 2.0 is 14.06, while there is a large long-tail distribution for cases that some users interact with customer service with relatively more turns.
Figure~\ref{fig3} shows the statistics of numbers of image. We can see that, in the most cases, customers only use one or two images in the conversation.
Figure~\ref{fig4} illustrates the statistics of dialogue utterance length. Users tend to use short sentences of about 20 characters to describe the problems they encounter, while customer assistant sometimes prefer longer sentences.


\begin{table}[!htb]
\centering
\scalebox{0.95}{
\begin{tabular}{|c|l|}
\hline
\textbf{Categories}     & \textbf{Sub-categories}                       \\ \hline
\multirow{9}{*}{Screenshot} & Screenshot of product                         \\ \cline{2-2} 
                            & Screenshot of product order                          \\ \cline{2-2} 
                            & Screenshot of logistics order                    \\ \cline{2-2} 
                            & Screenshot of after-sales service          \\ 
                            & order \\ \cline{2-2} 
                            & Screenshot of text message                    \\ \cline{2-2} 
                            & Screenshot of user comment                       \\ \cline{2-2} 
                            & Screenshot of system or software                \\ \cline{2-2} 
                            & Screenshot in other scenes            \\ \hline
\multirow{10}{*}{Photo}      & Photo for purchasing consultation         \\ \cline{2-2} 
                            & Photo of product with damaged \\
                            & appearance       \\ \cline{2-2} 
                            & Photo of products with malfunction            \\ \cline{2-2} 
                            & Photo of product with missing \\
                            & items            \\ \cline{2-2} 
                            & Photo for product recommendation \\
                            & and comparison \\ \cline{2-2} 
                            & Photo for product installation \\ \cline{2-2} 
                            & Photo of user screen shot                      \\ \hline
\end{tabular}
}
\caption{Definition of image categories and sub-categories.}
\label{tab.img}
\end{table}

\begin{figure*}[!htb]
\centering
\includegraphics[width=6.5 in]{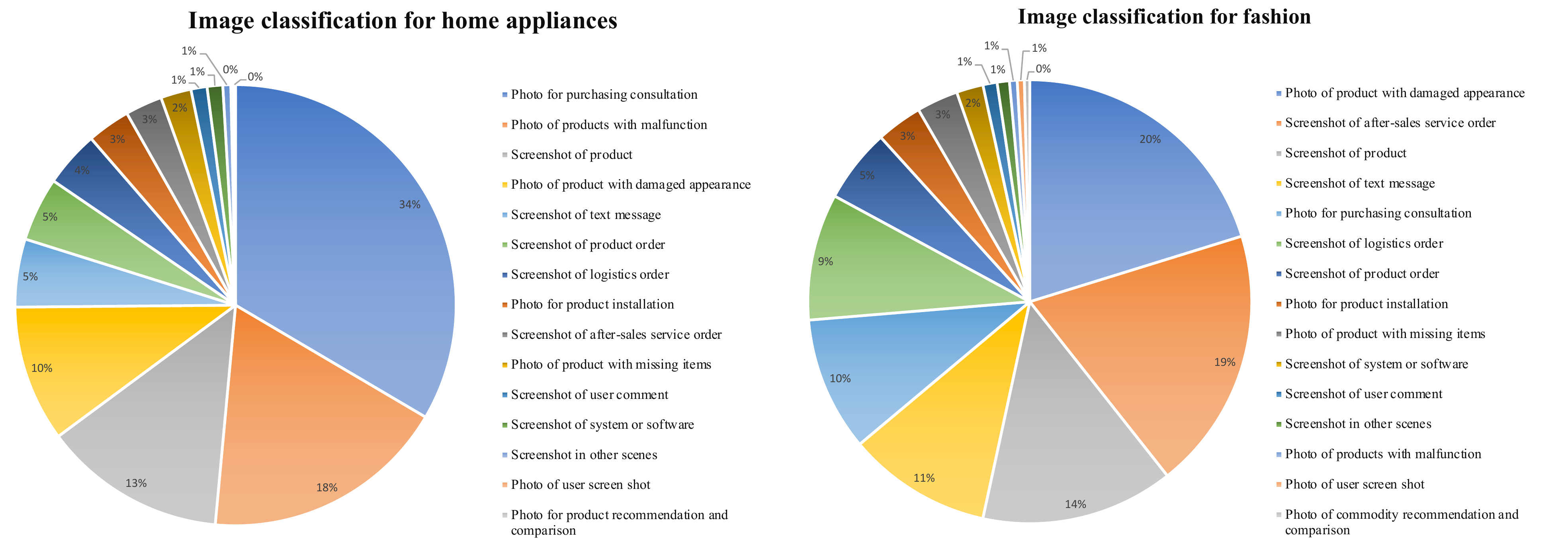}\\
\caption{Image classification for 15 sub-categories.}
\label{fig5}
\end{figure*}

\subsection{Image category}

The images in JDDC 2.0 can be divided into two categories: screenshot and real photo. Further, we classify these two categories into 15 sub-categories according to the characteristics of the business scenarios involved in the entire shopping process, such as pre-sales consultation, order payment, logistics distribution, and after-sale fault maintenance. Table~\ref{tab.img} shows the details of these sub-categories. We randomly select 5,000 dialogue sessions from the product category of small home appliances, of which the user questions contain 8,218 images, and 5,000 from fashion, of which the user questions contain 8,731 images. We annotate these images according to the sub-categories, which are also be released as part of our dataset. Figure~\ref{fig5} shows the statistical results. The top-3 sub-categories for small home appliances are:  34\% are photos for purchasing consultation, 18\% are photos of products with malfunction, 13\% are screenshots of product. 
The top-3 sub-categories for fashion are:  20\% are photos of product with damaged appearance, 19\% are screenshots of after-sales service order, 14\% are screenshots of product. 
The classification results show that, for small home appliances, the questions of user are mostly concentrated in pre-sales  consultation and after-sales product usage, while for fashion, users mainly concern about products with damaged appearance.

\begin{table*}[!htb]
\centering
\scalebox{1}{
\begin{tabular}{|l|c|l|}
\hline
\textbf{Item} & \textbf{Amount} & \textbf{Explanation}                                                             \\ \hline
Entity        & 30,205           & Products mentioned in the dialogue                                           \\ \hline
Entity type   & 231             & E.g., Rice cooker, air fryer, basketball shoes, jeans \\ \hline
Relation      & 759             & E.g., Anti-dry function, removable basket, upper material, waist type \\ \hline
Triple        & 219,121          & Detailed description of product attributes and product selling points\\\hline
\end{tabular}
}
\caption{The statistics of knowledge base.}
\label{tab.kb}
\end{table*}

\subsection{Knowledge Base}
In the E-commerce domain, whatever for the scenes of pre-sales purchasing consultation or after-sales return of product, the conversation always involves at least one product. 
Thus, for the product mentioned in the conversation, we provides the corresponding knowledge base that describes the attribute information in detail. 
The overall statistics of the knowledge base are shown in Table~\ref{tab.kb}. The knowledge base contains 30,205 products, involving 231 product sub-categories, and 759 types of product attribute. The total number of attribute triples is 219,121.

\section{Experiments}
In this section, we introduce the scheme and official results of JDDC 2021 dialogue challenge.  

\subsection{Task Definition}
The task of multimodal dialogue can be defined as:
\[ (H_{<n}, U_n, K \rightarrow R_n)   \]
where $H_{<n}$ denotes the historical dialogue before the $n$-th turn, $U_n$ denotes user question in the $n$-th turn, $K$ denotes product knowledge base information, $R_n$ denotes response in the $n$-th turn. That is, the task is to predict the the current response on condition of the given historical conversation records, a user question and the related knowledge base information. 

\subsection{JDDC 2021 Dialogue Challenge}
The JDDC 2.0 dataset has been divided into the training set, the validation set, and the test set. The training set and the validation set are available to the participating teams, while the test set is unseen. We take the first 5,000 dialogue sessions from the two product categories in the test set as the online test set. After the participating teams train the model offline, they submit the model to the online test platform, from which they can obtain evaluation results on the online test set.
In order to verify the ability of dialogue model to understand the visual information in dialogues, when we construct online test set,
we require the user questions in the $n$-th turn or in the previous turn must contain at least one image.

\subsection{Solutions of Top-5 Teams}

Team 1 combines the retrieval model and the generative model. For the retrieval model, they use rules to expand the knowledge base, screening out knowledge-aware question-answer pairs for question retrieval; for the generative model, they combine image categories, dialogue history, and external knowledge base as input and use Transformer~\cite{attn-17} to generate the responses. In addition, they use TextCNN~\cite{txtcnn-14} to build an intent recognition model. If the user intention is product information consultation or after-sales product quality, the response is given by the retrieval model, otherwise given by the generative model.

Team 2 applies the multi-modal generation model of hierarchical encoding and decoding based on GRU~\cite{gru-14}. The image features are extracted using ResNet~\cite{res-16}. The text and image are encoded separately and then merged into the context representation with two attention layers, among which the static attention layer is a self-attention layer for the encoded context and the dynamic attention layer filters the information when decoding the response at each time-step. 
Finally, the response is generated by the decoding layer.

Team 3 adopts a multi-modal fusion model based on GPT~\cite{gpt-18}, which puts the product knowledge triples corresponding to the product in the dialogue at the beginning of the input. It uses ResNet to extract the features of image and integrate them with text features by concatenation, and finally, GPT is used to generate the response.

Team 4 uses a multi-modal dialogue generation model based on pre-trained language models and structured knowledge bases. It first adopts domain adaptation pre-training, by masking the relationship and object entities in the product knowledge base, to train a dialogue model oriented to the structured knowledge base. Then, it uses the ResNet model to extract image features and K-means to perform an image clustering. The image is compressed into 200 types of token information that is then integrated with text token information into the previously trained dialogue model, which supports multi-modal information and knowledge information at the same time.

Team 5 adopts a single-modal dialogue generation model based on the UniLM model~\cite{unilm-19}. It uses JDAI-BERT-Chinese\footnote{The pre-trained models and embeddings are available at \url{https://github.com/jd-aig/nlp\_baai}.}, a large-scale pre-training BERT for E-commerce scenarios, as initial weights and trains UniLM to encode up to 5 turns of dialogue to generate the target response. 

\subsection{Experimental Results}
The results of the automatic and human evaluation are shown in Table~\ref{tab.bleu}. In this dialogue challenge, we use BLEU~\cite{bleu-Z02} as the evaluation criteria to rank the participating teams. In addition, we present distinct n-grams~\cite{dist-16} to evaluate the diversity of the generated responses, Greedy Matching~\cite{gm-12} to evaluate the relevance between posts and generated responses at the word level, Embedding Average~\cite{ea-16} and Vector Extrema~\cite{ve-14} at the sentence level. For human evaluation, we sample 1,000 instances from the test set and invite customer service experts to evaluate the responses generated by each participating team model. The responses that are logically correct, informative, and relevant to users' questions are judged as qualified. The human evaluation score is the proportion of qualified response.

\begin{table*}[!htb]
\scalebox{0.80}{
\begin{tabular}{|c|c|c|c|c|c|c|c|c|c|c|}
\hline
\textbf{Team} & \textbf{BLEU-1} & \textbf{BLEU-2} & \textbf{BLEU-3} & \textbf{BLEU-4} & \textbf{Dist-1} & \textbf{Dist-2} & \textbf{\begin{tabular}[c]{@{}c@{}}Greedy \\ matching\end{tabular}} & \textbf{\begin{tabular}[c]{@{}c@{}}Average \\ matching\end{tabular}} & \textbf{\begin{tabular}[c]{@{}c@{}}Extrema \\ matching\end{tabular}} & \textbf{\begin{tabular}[c]{@{}c@{}}Human \\ evaluation\end{tabular}} \\ \hline
Team 1        & 0.1759          & 0.1438          & 0.1257          & 0.115           & 0.0126          & 0.1314          & 0.557                                                               & 0.849                                                                & 0.8903                                                               & 0.389                                                                \\ \hline
Team 2        & 0.2101          & 0.1719          & 0.1519          & 0.1407          & 0.006           & 0.0595          & 0.5795                                                              & 0.8726                                                               & 0.92                                                                 & 0.413                                                                \\ \hline
Team 3        & 0.2103          & 0.1733          & 0.1547          & 0.14403         & 0.008           & 0.1021          & 0.5639                                                              & 0.8645                                                               & 0.9143                                                               & 0.418                                                                \\ \hline
Team 4        & 0.2491          & 0.2113          & 0.1895          & 0.1757          & 0.0092          & 0.1206          & 0.5993                                                              & 0.8693                                                               & 0.9091                                                               & 0.456                                                                \\ \hline
Team 5        & 0.1831          & 0.1448          & 0.124           & 0.1118          & 0.0119          & 0.1298          & 0.5664                                                              & 0.8574                                                               & 0.9047                                                               & 0.426                                                                \\ \hline
\end{tabular}
}
\caption{The automatic and human evaluation results on the online test set.}
\label{tab.bleu}
\end{table*}

\begin{figure*}[!htb]
\centering
\includegraphics[width=6 in]{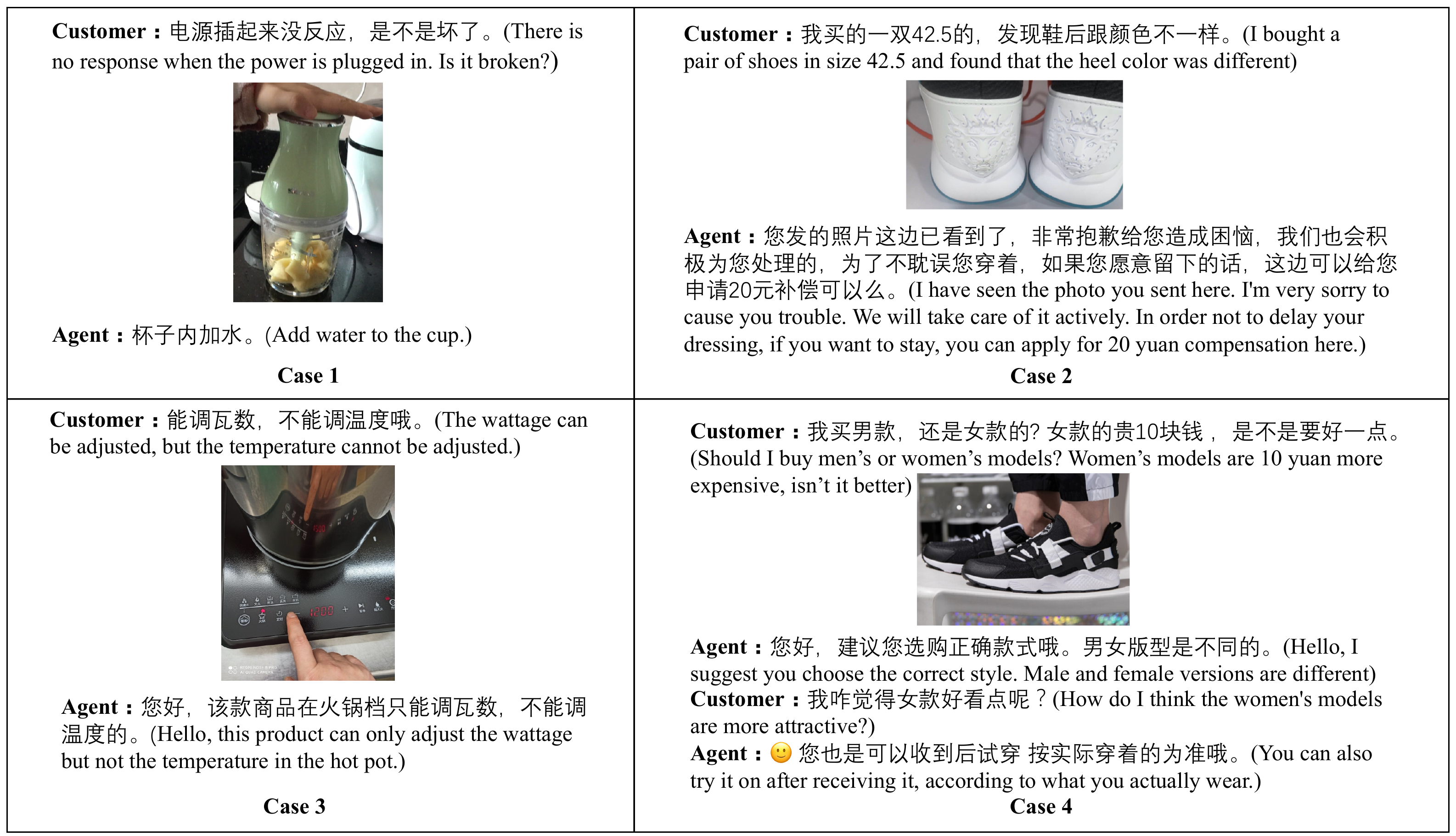}\\
\caption{Four typical examples in JDDC 2.0.}
\label{fig.case}
\end{figure*} 

\subsection{Case Study}
From the human evaluation results, we can conclude that the best model can only answer 45.6\% of the questions correctly, and thus, there are still many challenges to be solved in the multimodal dialogue task in the real E-commerce scenario. 
We show some cases in Figure~\ref{fig.case}.
For case 1, to generate a proper response, a dialogue model needs to recognize the product, i.e., a blender, in the image and understand the cause of product failures.
For case 2, there is a color difference in the product, but it does not affect usage. Generally, human customer service tends to compensate for it with money, while the models may be more inclined to return the product. This challenge requires the model to have more sophisticated dialogue strategies. 
For case 3, for the common question how to adjust the power of the induction cooker, almost all the models can give correct answers, while occasionally a few customers ask how to set the temperature on the induction cooker, most of the models fail to give proper responses. 
For case 4, it needs to effectively model the context to accurately express the detailed information such as the style favored by customers in the example.

\section{Privacy Protection}
Privacy is particularly important for a dataset from a real scenario. To protect the private information of merchants and customers in the conversation, we have designed some schemes to desensitize the data during the construction of the dataset:

(1) The store name, product brand, and model number are masked.

(2) The customer's name and address information are masked.

(3) The middle bits of the contact number are masked. It helped to hide more details of the speaker, such as phone numbers or order numbers, but keep useful short data, such as shoe size, cup capacity.

(4) The dialogue session number and product index are hashed to ensure that it cannot be recovered.

Moreover, we strictly restrict the use of the JDDC dataset to academic research, and any commercial product cannot use the dataset directly or indirectly. Anyone who downloads the data for our data site has to sign a agreement to obey these rules strictly.

\section{Conclusions and Future Work}
In this paper, we construct a real-scenario Chinese multi-modal dialogue dataset called JDDC 2.0 and provide part of the image annotation information and related product knowledge base information. We introduce the solutions of the top-5 teams in the JDDC 2021 challenge and give the official evaluation results on the online test set. These competition teams attempt to model the textual and visual information jointly and provide more satisfactory answers than with a single modality. However, there are still challenges in multimodal dialogue, such as understanding image details, effective modeling context information, long-tail questions, and refined dialogue strategy. The dialogue dataset is available for download at: \url{https://jddc.jd.com}. In the future, while exploring the multimodal joint modeling methods for real scene dialogue, we will provide more detailed annotation information on the dataset in terms of user emotions, user semantic analysis, and so on.

\section{Acknowledgements}
We would like to thank all participants of the JDDC 2021 dialogue challenge for providing multimodal dialogue solutions.
This work was supported by the National Key R\&D Program of China under Grant No. 2020AAA0108600.

\bibliography{custom}
\bibliographystyle{acl_natbib}

\end{document}